\documentclass[10pt]{article}

\usepackage[letterpaper,margin=0.82in]{geometry}
\usepackage[T1]{fontenc}
\usepackage[utf8]{inputenc}
\usepackage{lmodern}
\usepackage{microtype}
\usepackage{amsmath,amssymb,amsthm}
\usepackage{booktabs}
\usepackage{enumitem}
\usepackage{xcolor}
\usepackage{graphicx}
\usepackage{float}
\usepackage{tikz}
\usetikzlibrary{arrows.meta,positioning,fit}
\usepackage[
  colorlinks=true,
  linkcolor=black,
  citecolor=blue!55!black,
  urlcolor=blue!55!black,
  pdfencoding=auto,
  psdextra
]{hyperref}

\hypersetup{
  pdftitle={Jacobian-Causal Memory Control for Persistent Memory in Language Model Systems},
  pdfauthor={Marvin Danig},
  pdfsubject={Architecture and formal analysis preprint for controllable persistent AI memory},
  pdfkeywords={AI memory, language models, Jacobian-causal memory control, inference-time control, retrieval, rollback, agent memory}
}

\newtheorem{theorem}{Theorem}

\setlist{nosep,leftmargin=*}
\title{\textbf{Jacobian-Causal Memory Control for Persistent Memory\\in Language Model Systems}}
\author{Marvin Danig\\
\small Achiral Research\\
\small \href{mailto:marvin@achiral.ai}{\texttt{marvin@achiral.ai}}}
\date{\small Architecture and analysis preprint\\November 2026}

\begin{document}
\maketitle

\begin{abstract}
Persistent-memory systems choose not only which records are retrieved but which prior information can alter current inference. We propose Jacobian-causal memory control (JCMC), which represents each authorized memory intervention by a predicted influence signature in a shared behavioral measurement space, selects interventions against a target effect and side-effect budget, and compares predicted with observed effects before retaining or rolling back the intervention. Under explicit smoothness and bounded-estimation assumptions, we prove a quadratic local-approximation bound, characterize the target effects reachable by a set of signatures, bound selection regret under signature error, and bound non-additive interactions between simultaneous memories. The paper reports no benchmark measurements; it defines a paired counterfactual estimator and an evaluation protocol comparing JCMC with relevance, graph, and policy-filtered retrieval baselines. The central claim is narrowed or falsified if those simpler baselines match effect targeting, side-effect control, and rollback utility under equal model, memory, policy, and cost budgets.
\end{abstract}

\section{Introduction}

Long-lived language model systems now maintain state across conversations and tasks. Personal and agent memory systems retain preferences, project facts, files, summaries, tool outputs, and prior decisions, allowing future inference to start from accumulated context rather than an empty prompt \cite{openai2026dreaming,anthropic2026managedmemory,anthropic2025memory}. Open-weight frontier models are also increasingly optimized for agentic tool use and long-context work \cite{kimi2025k2}. These trends make memory a runtime control problem: the model does not merely retrieve information; it is shaped by what the system allows to become cognitively available. The motivating case is emergent memory, in which useful context forms through repeated work, reinforcement, forgetting, and revision rather than being stored by hand \cite{danig2026emergent}; the controller proposed here applies to persistent memory in general.

Retrieval-augmented generation (RAG) selects external information for generation \cite{lewis2020rag}. Graph, hierarchical, and agent-memory systems broaden the retrieval substrate and representation \cite{edge2024graphrag,sarthi2024raptor,packer2023memgpt,rasmussen2025zep}. Mechanistic interpretability and model-editing work shows that causal interventions on model internals can localize or alter behavior \cite{wang2022ioi,meng2022rome}. Influence functions trace training examples to predictions \cite{koh2017influence}. These literatures supply useful tools, but a persistent organizational memory system needs a distinct runtime question:
\begin{quote}
Which memory intervention should be permitted to affect this model now, what effect is expected, and what should happen if the effect is wrong?
\end{quote}

This paper proposes \emph{Jacobian-causal memory control} (JCMC) for persistent AI memory, named for the Jacobian-based local estimate of a memory's effect in Section~\ref{sec:signatures}. The central claim is narrow: persistent memory should be governed as a reversible, observed, lineage-bearing intervention rather than as a relevance-ranked context append. The claim is falsified or narrowed if simpler retrieval, graph-reranking, or prompt-policy baselines match the proposed controller on effect-targeting accuracy, side-effect control, and rollback utility under equal model, memory, policy, and cost budgets.

\paragraph{Contributions.}
We (i) distinguish memory relevance from memory influence (Section~\ref{sec:relevance}); (ii) specify influence signatures, constrained selection, paired observation, rollback, and lineage (Sections~\ref{sec:signatures}--\ref{sec:observation}); (iii) prove four conditional properties covering local approximation, reachability, selection regret, and multi-memory interaction (Section~\ref{sec:formal}); (iv) analyze implementation cost and controllability (Section~\ref{sec:cost}); and (v) provide a reproducible evaluation protocol (Section~\ref{sec:evaluation}). This architecture-and-analysis preprint reports no benchmark measurements.

\section{Memory Relevance Is Not Memory Influence}\label{sec:relevance}

Let $M=\{m_1,\ldots,m_n\}$ be a set of persistent memory entities and $q$ a current request. Conventional retrieval ranks candidates using a relevance score
\begin{equation}
s_q(m_i)=\operatorname{sim}(f(q),f(m_i)),
\end{equation}
where $f$ is an embedding function and $\operatorname{sim}$ a similarity measure, possibly combined with lexical, graph, metadata, or learned reranking features. Relevance answers whether a memory appears useful for the request. It does not measure the behavioral direction in which the memory will move a model once it is included.

For a compliance recommendation, a semantically relevant pricing memory may shift generation toward cost reduction when the intended effect is caution. A stale project summary may match the prompt while suppressing newer evidence. A high-confidence memory may become risky when combined with a particular user, tool, or policy state. In each case, the system needs an influence estimate, not only a relevance score.

We model a memory intervention as an operator $I(m_i,a_t)$ that applies memory $m_i$ through an application target $a_t$, such as the context window, a memory tool, an attention mechanism, or a latent-state interface. The intervention acts on a current model state and changes a future output distribution or a projected behavioral measurement $y$. The object of control is therefore
\begin{equation}
\Delta y_{future}(m_i,a_t \mid h_t,q,\pi),
\end{equation}
where $h_t$ is a model or task state and $\pi$ is the applicable authorization and policy regime.

\begin{table}[h]
\centering
\small
\begin{tabular}{p{0.22\linewidth}p{0.32\linewidth}p{0.34\linewidth}}
\toprule
\textbf{Dimension} & \textbf{Relevance retrieval} & \textbf{Jacobian-causal memory control} \\
\midrule
Primary question & What matches the request? & What will this memory cause the model to do? \\
Selection target & Passage, file, node, or summary & Reversible memory intervention \\
Evidence object & Similarity, metadata, graph proximity & Predicted and observed behavioral effect \\
Safety boundary & Filter before or after retrieval & Hard admissibility plus side-effect budget \\
Runtime feedback & Optional click, rating, or trace & Prediction, observation, divergence, lineage \\
Failure response & Drop, rerank, or ignore & Roll back, suppress, quarantine, refresh \\
\bottomrule
\end{tabular}
\caption{Relevance is a candidate signal. Jacobian-causal memory control treats recall as an intervention with expected and observed effects.}
\label{tab:distinction}
\end{table}

\section{Architecture}

The proposed controller contains six runtime components: a signature estimator, a signature store, an effect specification interface, an admissible-memory selector, an effect observer, and a lineage-governed rollback mechanism.

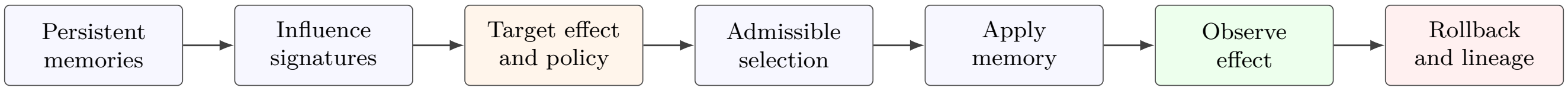
\begin{figure}[H]
\centering
\resizebox{\linewidth}{!}{%
\begin{tikzpicture}[
  box/.style={draw=black!70, rounded corners=2pt, align=center, minimum height=11mm, text width=22mm, fill=blue!3, font=\small},
  gate/.style={draw=black!70, rounded corners=2pt, align=center, minimum height=11mm, text width=22mm, fill=orange!8, font=\small},
  output/.style={draw=black!70, rounded corners=2pt, align=center, minimum height=11mm, text width=22mm, fill=green!7, font=\small},
  warn/.style={draw=black!70, rounded corners=2pt, align=center, minimum height=11mm, text width=22mm, fill=red!6, font=\small},
  flow/.style={-{Latex[length=2mm]}, line width=0.65pt, draw=black!75}
]
\node[box] (mem) {Persistent\\memories};
\node[box, right=7mm of mem] (sig) {Influence\\signatures};
\node[gate, right=7mm of sig] (spec) {Target effect\\and policy};
\node[box, right=7mm of spec] (select) {Admissible\\selection};
\node[box, right=7mm of select] (apply) {Apply\\memory};
\node[output, right=7mm of apply] (observe) {Observe\\effect};
\node[warn, right=7mm of observe] (diverge) {Rollback\\and lineage};

\draw[flow] (mem) -- (sig);
\draw[flow] (sig) -- (spec);
\draw[flow] (spec) -- (select);
\draw[flow] (select) -- (apply);
\draw[flow] (apply) -- (observe);
\draw[flow] (observe) -- (diverge);
\end{tikzpicture}}
\caption{Jacobian-causal memory control loop for persistent memory. The diagram omits product-specific coefficients, thresholds, and storage schemas; the research object is the observable control loop.}
\label{fig:architecture}
\end{figure}

\subsection{Model J-space}

Predicted and observed effects must be comparable. We call their shared representation \emph{model J-space}, a measurement space rather than a claim about the model's hidden state. It may be a full output-logit space, a task-specific token subspace, a sparse-feature space, a probe-score vector, or another projected measurement space. The choice is an engineering and evaluation decision. The publication-level requirement is that the system records the model version, prompt or probe protocol, projection, decoding policy, and measurement method used to produce each effect vector.

\subsection{Influence signatures}\label{sec:signatures}

For memory entity $m$, an influence signature $\phi_m$ estimates the directional shift caused by applying $m$ through a memory intervention operator. In a differentiable implementation, a local linearization can be written
\begin{equation}
\phi_m \approx P J_t M_m(c_m),
\end{equation}
where $c_m$ is memory content, $M_m$ maps the content into an intervention perturbation, $J_t$ is a future-output Jacobian or integrated influence map at state $h_t$, and $P$ projects the result into model J-space. When internal gradients are unavailable, $\phi_m$ can instead be estimated through finite differences, controlled probes, or learned local maps. This paper does not require one estimator to be universal.

A signature record should be treated as provisional unless it carries reliability evidence: model version, computation method, prompt or probe cohort, confidence intervals or residuals, staleness status, and calibration history. The record belongs to the memory governance layer, not to the memory text alone.

\subsection{Effect specification and selection}\label{sec:selection}

At runtime, a requester or policy process supplies a cognitive-effect specification containing a target influence vector $\phi_{\mathrm{target}}$, prohibited regions, side-effect budget, policy constraints, and intervention criticality. The selector evaluates only authorized candidates and may minimize
\begin{equation}
\mathcal{L}(m)=
\|\phi_m-\phi_{\mathrm{target}}\|^2
+\lambda\|\operatorname{proj}_{\perp}(\phi_m)\|^2
+\mu R_{\pi}(m)
+\nu U(m),
\end{equation}
where $\operatorname{proj}_{\perp}(\phi_m)$ is the component of $\phi_m$ orthogonal to $\phi_{\mathrm{target}}$ (the off-target influence), $R_{\pi}(m)$ is policy risk, $U(m)$ is signature uncertainty, and $\lambda,\mu,\nu\ge 0$ are weights set by the deployment. The formula is illustrative of the control objective: target the intended shift, penalize off-target influence, respect policy, and prefer reliable signatures. Candidates that violate hard access, provenance, or prohibited-region constraints are excluded rather than softly downweighted.

\subsection{Observation and rollback}\label{sec:observation}

After a memory intervention is applied, an observer estimates its effect through a paired counterfactual contrast. Let $z(h,\delta,\xi)\in\mathbb{R}^{r}$ be an output statistic under runtime state $h$, intervention coordinate $\delta$, and decoding randomness $\xi$, and let $P\in\mathbb{R}^{d\times r}$ project that statistic into model J-space. For $R$ paired runs using the same seeds or sampling draws in each intervention--baseline pair, define
\begin{equation}
\widehat{\Delta}_m=
\frac{1}{R}\sum_{j=1}^{R}P\left[z(h,\delta_m,\xi_j)-z(h,0,\xi_j)\right].
\end{equation}
Here $0$ denotes the matched run without the memory intervention. Shared randomness reduces decoding variance; deterministic decoding is the special case $R=1$. When shared randomness is unavailable, independent runs require uncertainty intervals wide enough to include sampling variance. The controller compares the paired estimate with the predicted signature using
\begin{equation}
D(m)=\frac{\|\widehat{\Delta}_m-\phi_m\|^2}{\|\phi_m\|^2+\epsilon_0},
\end{equation}
where $\epsilon_0>0$ keeps the ratio defined when $\phi_m$ is near zero. If $D(m)$ exceeds a threshold determined by criticality and uncertainty, the controller executes a response appropriate to the application target: remove context, revoke a memory-tool result, discard an attention bias, restore a checkpoint, suppress future selection, or quarantine the memory pending review. Rollback cannot undo effects that have already crossed the model boundary; it governs subsequent state and action.

\section{Formal Properties}\label{sec:formal}

\subsection{Notation and assumptions}

Fix a model version and runtime state $h$, including the prompt, conversation history, tool state, policy state, and decoding configuration. Let $\delta\in\mathbb{R}^{p}$ be an intervention coordinate, with $\delta=0$ denoting no memory intervention and $\delta_m=M_m(c_m)$ denoting the coordinate induced by memory content $c_m$. Let $z(h,\delta,\xi)\in\mathbb{R}^{r}$ be an integrable output statistic and $P\in\mathbb{R}^{d\times r}$ a fixed projection into model J-space. Define
\begin{equation}
F_h(\delta)=\mathbb{E}_{\xi}\left[Pz(h,\delta,\xi)\right],
\qquad
\Delta_h(\delta)=F_h(\delta)-F_h(0).
\end{equation}
For deterministic decoding, the expectation contains one outcome. For stochastic decoding, it is taken over a declared sampling distribution. The Jacobian is $J_h(\delta)=DF_h(\delta)\in\mathbb{R}^{d\times p}$, and the first-order influence signature is $\phi_h(\delta)=J_h(0)\delta$; for memory $m$, write $\phi_m=\phi_h(\delta_m)$. Because $F_h$ already includes the projection $P$, this notation is equivalent to the projected form $PJ_tM_m(c_m)$ in Section~\ref{sec:signatures}. Vector norms are Euclidean; matrix norms and bilinear-map norms are the induced operator norms.

The results below use only the assumptions named in each statement. In all cases, the model version and state $h$ are held fixed, intervention and baseline runs differ only through $\delta$, and the projection $P$ is unchanged. The term \emph{causal} refers to this controlled intervention contrast. It does not identify causal effects from uncontrolled observational logs.

\begin{theorem}[Local influence approximation]\label{thm:local}
Suppose $F_h$ is continuously differentiable on the line segment $\{t\delta:0\leq t\leq1\}$ and its Jacobian is $L$-Lipschitz there:
\begin{equation}
\|J_h(u)-J_h(v)\|\leq L\|u-v\|.
\end{equation}
Then the first-order signature satisfies
\begin{equation}
\|\Delta_h(\delta)-\phi_h(\delta)\|
\leq \frac{L}{2}\|\delta\|^2.
\end{equation}
\end{theorem}

\begin{proof}
The fundamental theorem of calculus gives
\begin{equation}
\Delta_h(\delta)-J_h(0)\delta
=\int_0^1\left[J_h(t\delta)-J_h(0)\right]\delta\,dt.
\end{equation}
Taking norms and applying the Lipschitz assumption yields
\begin{equation}
\|\Delta_h(\delta)-J_h(0)\delta\|
\leq\int_0^1 Lt\|\delta\|^2dt
=\frac{L}{2}\|\delta\|^2.
\end{equation}
\end{proof}

Theorem~\ref{thm:local} justifies a local signature only when the intervention is small relative to the curvature of $F_h$. It does not justify transporting the signature across model versions or substantially different runtime states.

\begin{theorem}[Reachable target effects]\label{thm:reachability}
Let $\Phi=[\phi_1,\ldots,\phi_T]\in\mathbb{R}^{d\times T}$ collect $T$ influence signatures, let $\tau\in\mathbb{R}^{d}$ be a target effect, and let $P_{\Phi}=\Phi\Phi^{\dagger}$ be the orthogonal projector onto the column space of $\Phi$, where $\Phi^{\dagger}$ is the Moore--Penrose pseudoinverse. Under the first-order additive model, unrestricted coefficients $\alpha\in\mathbb{R}^{T}$ satisfy
\begin{equation}
\min_{\alpha}\|\Phi\alpha-\tau\|^2
=\|(I_d-P_{\Phi})\tau\|^2.
\end{equation}
The target is exactly reachable if and only if $\tau$ lies in the column space of $\Phi$.
\end{theorem}

\begin{proof}
Decompose $\tau=P_{\Phi}\tau+(I_d-P_{\Phi})\tau$. The two terms are orthogonal, every vector $\Phi\alpha$ lies in the column space of $\Phi$, and $P_{\Phi}\tau$ is the unique closest point in that subspace. The residual is therefore $(I_d-P_{\Phi})\tau$, which vanishes exactly when $\tau$ belongs to the column space.
\end{proof}

Authorization, sign, sparsity, or magnitude constraints on $\alpha$ can only increase this minimum error. The theorem is therefore an optimistic lower bound for a constrained deployment, not evidence that arbitrary effects should be pursued.

\begin{theorem}[Selection regret under bounded estimation error]\label{thm:regret}
Let $\mathcal{A}$ be a finite set of authorized memory interventions. Let $\mathcal{L}(m)$ be the true deployment loss and $\widehat{\mathcal{L}}(m)$ its signature-based estimate. Suppose
\begin{equation}
|\widehat{\mathcal{L}}(m)-\mathcal{L}(m)|\leq\varepsilon
\quad\text{for every }m\in\mathcal{A}.
\end{equation}
If $\widehat{m}\in\arg\min_{m\in\mathcal{A}}\widehat{\mathcal{L}}(m)$ and $m^{\star}\in\arg\min_{m\in\mathcal{A}}\mathcal{L}(m)$, then
\begin{equation}
\mathcal{L}(\widehat{m})
\leq\mathcal{L}(m^{\star})+2\varepsilon.
\end{equation}
\end{theorem}

\begin{proof}
By the uniform error bound and the optimality of $\widehat{m}$ under the estimated loss,
\begin{align}
\mathcal{L}(\widehat{m})
&\leq\widehat{\mathcal{L}}(\widehat{m})+\varepsilon\\
&\leq\widehat{\mathcal{L}}(m^{\star})+\varepsilon\\
&\leq\mathcal{L}(m^{\star})+2\varepsilon.
\end{align}
\end{proof}

If the effect-dependent part of the loss is $K$-Lipschitz and every estimated signature is within $\rho$ of its corresponding expected effect, while all other loss terms are exact, then $\varepsilon\leq K\rho$ and the regret bound becomes $2K\rho$. The guarantee is only as strong as the uniform calibration bound.

\begin{theorem}[Bound on multi-memory interaction]\label{thm:interaction}
Suppose $F_h$ is twice continuously differentiable on the parallelogram $\{s\delta_1+t\delta_2:0\leq s,t\leq1\}$ and its second derivative satisfies
\begin{equation}
\|D^2F_h(x)[u,v]\|\leq H\|u\|\|v\|
\end{equation}
for all points $x$ in that set and all $u,v\in\mathbb{R}^{p}$. Define the non-additive interaction
\begin{equation}
C_h(\delta_1,\delta_2)
=F_h(\delta_1+\delta_2)-F_h(\delta_1)-F_h(\delta_2)+F_h(0).
\end{equation}
Then
\begin{equation}
\|C_h(\delta_1,\delta_2)\|
\leq H\|\delta_1\|\|\delta_2\|.
\end{equation}
\end{theorem}

\begin{proof}
Applying the fundamental theorem of calculus along both intervention coordinates gives
\begin{equation}
C_h(\delta_1,\delta_2)
=\int_0^1\int_0^1
D^2F_h(s\delta_1+t\delta_2)[\delta_1,\delta_2]\,dt\,ds.
\end{equation}
The stated bound follows by taking norms and applying the uniform Hessian bound over the integration region.
\end{proof}

Theorem~\ref{thm:interaction} states when independent signatures may be approximately composed. Large curvature, large intervention norms, or a failure of smoothness invalidates the additive approximation and requires direct joint measurement. These four theorems are conditional properties of the specification; none establishes empirical advantage over a retrieval baseline.

\section{Analytic Consequences}\label{sec:consequences}

\paragraph{C1: The retrieval unit changes.}
The selected object is no longer simply the most relevant document or passage. It is an admissible intervention expected to move model behavior in a desired direction.

\paragraph{C2: Side effects become measurable design objects.}
A memory can be on-topic and still harmful because it pushes the model toward an unwanted frame. JCMC makes off-target projection a first-class cost.

\paragraph{C3: Reliability is memory-specific and model-version-specific.}
A signature estimated for one checkpoint, prompt protocol, or tool interface may become stale when the model, decoding policy, or memory application target changes.

\paragraph{C4: Memory safety includes reversibility.}
Access control prevents unauthorized use; reversibility addresses authorized memories whose effects are inaccurate, stale, adversarial, or contextually unsafe.

\paragraph{C5: Lineage converts failures into calibration data.}
Every confirmed or rolled-back intervention provides evidence about estimator quality, memory staleness, multi-memory interaction, and policy adequacy.

\section{Cost and Controllability}\label{sec:cost}

Influence estimation is more expensive than ordinary retrieval. A practical system therefore separates offline signature preparation, cheap candidate filtering, and high-fidelity validation for sensitive contexts. Full Jacobians are usually impractical at runtime for frontier-scale models; integrated gradients, low-rank approximations, probe-based estimates, and cached signatures are more plausible operational regimes.

For $N$ memory entities and influence dimension $d$, dense signature storage costs $O(Nd)$ before compression. Candidate selection can be reduced using approximate nearest-neighbor search over signatures after authorization and metadata filters. Observed-effect measurement costs depend on probe count, output horizon, and whether the system can inspect logits or only scored responses.

Control-theoretic diagnostics are useful when planning multiple interventions. For a plan of $T$ interventions $m_1,\ldots,m_T$, a reachability matrix or controllability Gramian over their signatures can identify target directions poorly reachable by memory injection alone:
\begin{equation}
W_c=\sum_{k=1}^{T}\phi_{m_k}\phi_{m_k}^{\top}.
\end{equation}
Small eigenvalue directions should be treated cautiously: failure to reach them may reflect the limits of the memory substrate rather than a selection bug. An observability analogue can test whether the proposed probe set can reliably distinguish the intended influence direction from noise.

\section{Evaluation and Falsification}\label{sec:evaluation}

The proposed architecture should be evaluated against simpler baselines before deployment claims are made. We propose matched comparisons among:
\begin{enumerate}
\item dense or hybrid top-$k$ retrieval;
\item graph retrieval or memory-tool recall without influence signatures;
\item relevance retrieval plus policy filtering;
\item influence reranking without observation;
\item full JCMC with observation, rollback, reliability, and lineage.
\end{enumerate}

Tasks should include compliance-sensitive recommendations, temporal supersession, project-memory recall, contradiction handling, tool-use planning, multi-session agent work, and adversarial or stale memory injection. Primary metrics should include target-effect alignment, side-effect magnitude, answer accuracy, provenance completeness, stale-memory false-positive rate, rollback precision, post-rollback recovery quality, calibration error, latency, token cost, and human audit time.

The central claim is narrowed if relevance or graph baselines match target-effect alignment and side-effect control at lower cost. It is falsified if observed-effect measurement fails to predictably distinguish beneficial interventions from harmful ones, or if rollback and lineage do not improve post-failure recovery.

\paragraph{Reproducibility protocol.}
A future empirical paper should publish or escrow the corpus snapshot or synthetic generator, memory schema, authorization rules, signature-estimation protocol, projection definition, model and decoding versions, random seeds, probe templates, per-task traces, selected and rejected memories, observed effects, rollback events, and aggregate plus per-stratum metrics. This preprint releases no benchmark dataset, production thresholds, or product implementation.

\paragraph{Prospective evaluation infrastructure.}
The protocol could be implemented as a dedicated JCMC study within \href{https://github.com/Achiral-AI/cognoscenti}{Cognoscenti}, an open benchmark framework for AI memory systems. Such a study requires new intervention-level instrumentation for paired counterfactual runs, predicted and observed influence signatures, divergence detection, rollback, and post-rollback recovery. No JCMC evaluation has yet been conducted using Cognoscenti; the workload, implementation, and results remain future work.

\section{Limitations, Failure Modes, and Scope}

\paragraph{Limits of this paper.} The paper reports no measurements, so every benefit claimed for JCMC is a hypothesis until the evaluation in Section~\ref{sec:evaluation} is run. The Jacobian estimate in Section~\ref{sec:signatures} is local: it describes small perturbations around the state $h_t$ and may mislead when a memory changes the model's behavior sharply, as when a single recalled policy flips a recommendation from approve to decline. For closed-weight models served through an API, gradients are unavailable, so signatures must come from finite differences or probes, which cost more and are noisier.

The formal results are conditional. Theorems~\ref{thm:local} and~\ref{thm:interaction} require local smoothness; Theorem~\ref{thm:reachability} uses a first-order additive model with unrestricted coefficients; and Theorem~\ref{thm:regret} assumes a uniform loss-estimation bound that must itself be measured. None of these assumptions establishes that influence signatures are accurate for a deployed model.

\paragraph{Failure modes.} JCMC also creates its own risks:
\begin{itemize}
\item \emph{Signature leakage.} Stored signatures may reveal private memories if not protected. For example, a signature that reliably shifts answers toward a particular vendor can disclose that a confidential negotiation with that vendor exists.
\item \emph{Probe overfitting.} Probe protocols may overfit to narrow behavior. A probe set built from compliance questions may report a pricing memory as harmless while it still shifts tone in sales conversations the probes never cover.
\item \emph{Incomplete rollback.} Rollback cannot recall what has already left the model boundary. If an agent has already sent an email drafted under a stale project summary, removing the summary only protects future answers.
\item \emph{Non-composition.} Multi-memory interactions can be nonlinear, making single-memory signatures unreliable. Two memories that are each benign, a deadline and a budget cut, may together push the model toward skipping a safety review that neither would trigger alone.
\item \emph{Adversarial memories.} Attackers may craft memories with benign surface meaning and harmful influence signatures, such as a routine-looking meeting note written to make later answers dismiss a security alert.
\item \emph{Manipulation.} A controller optimized for target effect can become manipulative if targets are not constrained by user intent, organizational policy, and law, for instance by selecting memories that steer a user toward an upsell rather than the answer they asked for.
\end{itemize}

The intended scope is not short-lived factual question answering over a small corpus. The architecture is most relevant for long-running agents, enterprise memory, compliance-sensitive work, model-assisted decision processes, and settings where memory influence must be attributable and reversible.

\section{Related Work and Claim Boundary}

RAG and its graph or hierarchical variants improve the selection and organization of external information for generation \cite{lewis2020rag,edge2024graphrag,sarthi2024raptor}. Agent-memory systems such as MemGPT, temporal knowledge-graph memory, Claude memory features, and ChatGPT memory systems address persistence, summarization, and continuity across sessions \cite{packer2023memgpt,rasmussen2025zep,anthropic2026managedmemory,openai2026dreaming}. Mechanistic interpretability and model-editing work studies causal circuits and interventions within models \cite{wang2022ioi,meng2022rome}. Influence functions attribute predictions to training examples \cite{koh2017influence}. Recent interpretability work on global workspace-like structures suggests that modern language models may expose controllable internal patterns relevant to deliberate reasoning \cite{anthropic2026workspace}.

Two recent systems independently test parts of this paper's motivating argument empirically. Srivastava reframes memory selection as an interventional decision problem and reports that selecting by estimated causal contribution outperforms vector (0.839), graph (0.824), and summary (0.723) retrieval on a causal question-answering benchmark (task score 0.846), though vector retrieval edges it out on temporal sub-questions specifically \cite{srivastava2026causal}; this is direct empirical evidence for the distinction drawn in Section~\ref{sec:relevance} between relevance and influence. The estimator there is a black-box ablation over three forward passes scored against a fixed threshold, not a stored per-memory signature, and selection is not posed as matching a target effect under a side-effect budget; there is no post-application observation step and no rollback. Yu et al. report that dependency-graph-guided rollback recovers substantially more accuracy (0.853) than full-context reset (0.407) or naive deletion (0.340) after a memory fault \cite{yu2026rollback}, independent empirical support for this paper's claim that reversibility is only useful if it is targeted (Section~\ref{sec:observation}). Their rollback is triggered by a diagnosis supplied by an external detector after a failure is already observed, not by comparing a pre-registered prediction against a paired post-application observation, and selecting which memory to apply in the first place is out of scope for that work. Neither system stores a causal signature per memory before application or verifies it against an observed effect afterward, which is the specific loop this paper proposes.

This paper does not claim those components independently. The proposed contribution is the ordered runtime architecture: persistent memory entities carry influence signatures; a controller selects interventions by target effect, side-effect budget, policy risk, and uncertainty; observed effects are compared to predicted effects; and divergence triggers rollback, suppression, quarantine, and lineage updates.

\section{Conclusion}

Persistent memory changes the status of retrieval: a recalled memory is an intervention that can alter future model behavior. JCMC specifies how to predict that influence, constrain side effects, compare prediction with a paired observed effect, and preserve lineage when prediction fails. Under the stated assumptions, the formal results bound local approximation error, characterize first-order reachability, limit selection regret, and quantify pairwise interaction. They do not establish empirical advantage. That question requires the matched evaluation defined in Section~\ref{sec:evaluation}.

\section*{Declarations}

\paragraph{Author contributions.}
Marvin Danig conceived the architecture, developed the analytic specification, and wrote the manuscript.

\paragraph{Funding.}
This research received no external funding.

\paragraph{Patent and rights notice.}
The work is associated with Achiral Research and U.S. provisional patent application materials concerning Jacobian-causal memory control for persistent AI memory systems. Publication of this preprint does not grant any patent license.

\paragraph{Data, code, and materials availability.}
This paper reports an architectural specification, formal analysis, and an evaluation protocol; it introduces no experimental dataset or benchmark code. A companion research essay and public preprint links are maintained at
\href{https://achiral.ai/research/jacobian-causal-memory-control}{\texttt{achiral.ai/research/jacobian-causal-memory-control}} and
\href{https://achiral.ai/papers}{\texttt{achiral.ai/papers}}.

\paragraph{Ethics and limitations.}
The proposed system could be misused to manipulate model behavior through persistent memory. Deployment requires user-intent constraints, authorization before scoring, privacy protection for signatures, robust audit logs, calibrated rollback, and human review for high-stakes memory effects.

\end{document}